\documentclass[sigconf]{acmart}

\usepackage{booktabs} 
\usepackage{multicol}
\usepackage{balance} 
\setcopyright{acmcopyright}
\usepackage{multirow}

\copyrightyear{2018}
\acmYear{2018}
\setcopyright{acmcopyright}
\acmConference[ICMI '18]{2018 International Conference on Multimodal
Interaction}{October 16--20, 2018}{Boulder, CO, USA}
\acmBooktitle{2018 International Conference on Multimodal Interaction (ICMI
'18), October 16--20, 2018, Boulder, CO, USA}
\acmPrice{15.00}
\acmDOI{10.1145/3242969.3264972}
\acmISBN{978-1-4503-5692-3/18/10}

\begin{document}


\title{EmotiW 2018: Audio-Video, Student Engagement and Group-Level Affect Prediction}

\author{ Abhinav Dhall }
\affiliation{Indian Institute of Technology Ropar, India}
\email{abhinav@iitrpr.ac.in}

\author{ Amanjot Kaur }
\affiliation{Indian Institute of Technology Ropar, India}
\email{amanjot.kaur@iitrpr.ac.in}

\author{ Roland Goecke }
\affiliation{University of Canberra, Australia}
\email{roland.goecke@ieee.org}

\author{Tom Gedeon}
\affiliation{Australian National University, Australia}
\email{tom.gedeon@anu.edu.au}

\begin{abstract}
This paper details the sixth Emotion Recognition in the Wild (EmotiW) challenge. EmotiW 2018 is a grand challenge in the ACM International Conference on Multimodal Interaction
2018, Colarado, USA. The challenge aims at providing a common platform to researchers working in the affective computing community to benchmark their algorithms on
`in the wild' data. This year EmotiW contains three sub-challenges: a) Audio-video based emotion recognition; b) Student engagement prediction; and c) Group-level emotion recognition. The databases, protocols and baselines are discussed in detail.

\end{abstract}


\maketitle
\section{Introduction}

\begin{figure}[t]
		\begin{center}
			\includegraphics[width=86mm]{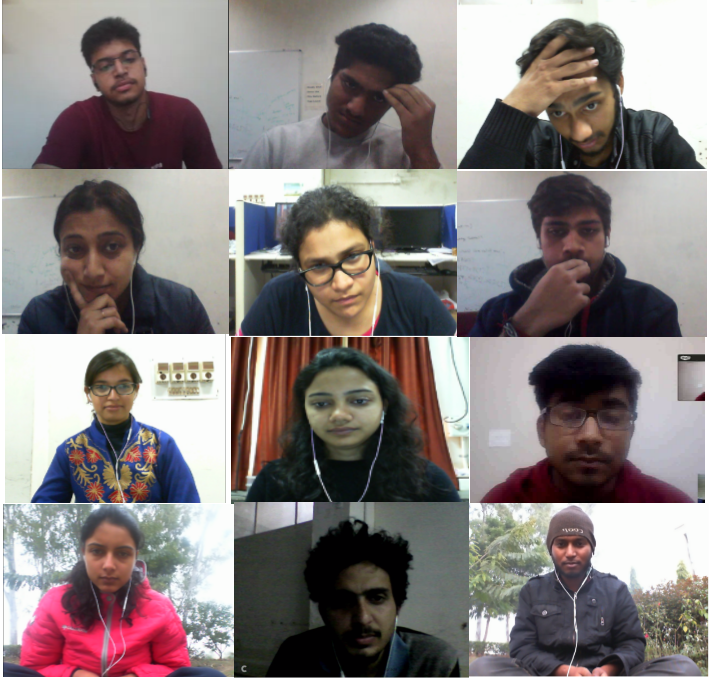}
			\caption{The images of the videos in the student engagement recognition sub-challenge \cite{mustafa2018prediction}. Please note the varied backgrounds environment and illumination.}
			\label{fig:GReco}
		\end{center}
	\end{figure}

The sixth Emotion Recognition in the Wild (EmotiW)\footnote{\url{https://sites.google.com/view/emotiw2018}} challenge is a series of benchmarking effort focussing on different problems in affective computing in real-world environments. This year's EmotiW is part of the ACM International Conference on Multimodal Interaction (ICMI) 2018. EmotiW is a challenge series annually organised as a grand challenge in ICMI conferences. The aim is to provide a competing platform for researchers in affective computing. For details about the earlier EmotiW challenge, please refer to EmotiW 2017's baseline paper \cite{dhall2017individual}. There are other efforts in the affective computing community, which focus on different problems such as depression analysis (Audio/Video Emotion Challenge \cite{ringeval2017avec}) and continous emotion recognition (Facial Expression Recognition and Analysis \cite{valstar2017fera}). Our focus is affective computing in `in the wild' environments. Here `in the wild' means different real-world conditions, where subjects show head pose change, have varied illumination on the face, show spontaneous facial expression, there is background noise and occlusion etc. An example of the data captured in different environments can be seen in Figure \ref{fig:GReco}.

EmotiW 2018 contains three sub-challenges: a) Student Engagement Prediction (EngReco); Audio-Video Emotion Recognition (VReco); and c) Group-level Emotion Recognition (GReco). EngReco is a new problem introduced this year. In total there were over 100 registrations in the challenge. Below, we discuss the three sub-challenges, their baseline, data, evaluation protocols and results.

\section{Student Engagement Recognition}
Student engagement in MOOCs is a challenging task. Engagement is one of the affective state which is a link between the subject and resource. It has various aspects such as emotional, cognitive and behavioral aspect. Challenge involved in engagement level detection of a user is that it does not remain same always, while watching MOOC material. To help the students to retain their attention level or track those parts of the video where they loss the attention it is mandatory to track student engagement based on various social cues such as looking away from the screen, feeling drowsy, yawning, being restless in the chair and so on.
User engagement tracking is vital for other application such as detecting vehicle driver \textquotesingle s attention level while driving, customer engagement while reviewing new product. With the advent of e-learning environment in the education domain automatic detection of engagement level of students based on computer vision and machine learning technologies is the need of the hour. An algorithmic approach for automatic detection of engagement level requires dataset of student engagement. Due to unavailability of datasets for student engagement detection in the wild new dataset for student engagement detection is created in this work.
It will address the issue of creating automatic student engagement tracking software. It will be used for setting performance evaluation benchmark for student engagement detection algorithms. In the literature, various experiments are conducted for student engagement detection in constrained environment. Various features used for engagement detection are based on Action Units, facial landmark points, eye movement, mouse clicks and motion of head and body.

\subsection{Data Collection and Baseline}
Our database collection details are discussed in the Kaur et al.\ \cite{mustafa2018prediction}. Student participants were asked to watch five minutes long MOOC video. The data recording was done with different methods: through Skype, using a webcam on a laptop or computer and using a mobile phone camera. We endeavored to capture data in different scenarios. This is inorder to simulate different environments, in which students watch learning materials. The different environments used during the recording are computer lab, playground, canteen, hostel rooms etc. In order to introduce unconstrained environment effect different lighting conditions are also used as dataset is recorded at different times of the day. Figure \ref{fig:GReco} shows the different environments represented in the engagement database.

The data was divided into three sub-sets: \emph{Train}, \emph{Validation} and \emph{Test}. In the dataset total 149 videos for training and 48 videos for validation are released. Testing data contains 67 videos. Dataset split follows subject independence i.e no subject is repeated among the three splits. The class wise distribution of data is as follows: 9 videos belong to level 0, 45 videos belong to level 1, 100 videos belong to level 2 and remaining 43 videos belong to level 3. The dataset has total 91 subjects (27 females and 64 males) in total. The age range of the subjects is 19-27 years. The annotation of the dataset is done by 6 annotators. The inter reliability of the annotators is measured using weighted Cohen\textquotesingle s $K$ with quadratic weights as the performance metric. This is a regression problem so the labels are in the range of [0 - 1].    

For the baseline eye gaze and head movement features are computed. The eye gaze points and head movement w.r.t to camera movement are extracted using the OpenFace library \cite{baltruvsaitis2016openface}. The approach is as follows: firstly, all the videos are down sampled to same number of frames. The video is then divided into segments with 25\% overlap. For each segment statistical features are generated such as standard deviation of the 9 features (from OpenFace). As a result each video has 100 segments, where each segment is represented with the help of the 9 features. By learning a long short term memory network the Mean Square Error (MSE) are 0.10 and 0.15 for the \emph{Validation} and the \emph{Test} sets, respectively. The performance of the competing teams on the \emph{Test} set can be viewed in Table \ref{tab:Engagement}. A total of 6 teams submitted the labels for evaluation during the testing phase. Please note that this list is preliminary as the evaluation of code of the top three teams is underway. The same applies to the other two sub-challenges.  

\section{Group-level Emotion Recognition}

This sub-challenge is the continuation of EmotiW 2017's GReco sub-challenge \cite{dhall2017individual}. The primary motivation behind this is to be able to predict the emotion/mood of a group of people. Given the large increase in the number of images and videos, which are posted on social networking platforms, there is an opportunity to analyze affect conveyed by a group of people. The task of the sub-challenge is to classify a group's perceived emotion as \emph{Positive}, \emph{Neutral} or \emph{Negative}. The labeling is representation of the Valence axis. The images in this sub-challenge are from the Group Affect Database 3.0 \cite{dhall2015more}. The data is distributed into three sets:  \emph{Train}, \emph{Validation} and \emph{Test}. The \emph{Train}, \emph{Validation} and \emph{Test} sets contain 9815, 4346 and 3011 images, respectively. As compared to the EmotiW 2017 the amount of data has increased three folds.

For computing the baseline, we trained the Inception V3 network followed by three fully connected layers (each having 4096 nodes) for the three classification task. We use stochastic gradient descent optimizer without any learning rate decay to train the model. The classification accuracy for the \emph{Validation} and \emph{Test} sets are 65.00\% and 61.00\%, respectively. The performance of the competing teams in this sub-challenge are reported in the Table \ref{tab:Group}. A total of 12 teams submitted labels for evaluation during the testing phase.

\section{Audio-video based Emotion Recognition}
The VReco sub-challenge is the oldest running task in the EmotiW challenge series. The task is based on the Acted Facial Expressions in the Wild (AFEW) database \cite{Dhall_MM_2012}. AFEW database has been collected from movies and TV serials using a keyword search. Subtitles for hearing impaired contain keywords, which may correspond to the emotion of the scene. The short sequences with subtitles containing emotion related words were used as candidate samples. The database is then curated with these candidate audio-video samples. The database similar to the other two databases in EmotiW has been divided into three subsets: \emph{Train}, \emph{Validation} and \emph{Test}.

The task is to predict the emotion of the subject in the video. Universal categorical emotion representation (\emph{Angry}, \emph{Disgust}, \emph{Fear}, \emph{Happy}, \emph{Neutral}, \emph{Sad} and \emph{Surprise}) is used for representing emotions.

The baseline is computed as follows: face detection \cite{RamananCVPR2012} is performed for initializing the tracker \cite{XiongTorre2013}. The face volume of aligned faces is divided into non-overlapping patches of $4\times4$ and Local Binary Patterns in Three Orthogonal Planes (LBP-TOP)\cite{Guo_Dynamic_Texture_PAMI_2007} is computed. LBP-TOP captures the spatio-temporal changes in the texture. For classification, we trained a non-lines support vector machine. The classification accuracy (\%) on the \emph{Validation} and \emph{Test} set are 38.81\% and 41.07\%, respectively. The data in this sub-challenge is similar to that of EmotiW 2017 \cite{dhall2017individual}. Table \ref{tab:AV} shows the comparison of the classification accuracy for 31 teams in this sub-challenge. It is notable that the performance of most the teams outperforms the baseline. Most of the proposed techniques are based on deep learning.

\begin{table}[]
\begin{tabular}{|l|l|c|}
\hline
\multicolumn{3}{|c|}{\multirow{2}{*}{\textbf{Engagement Prediction Challenge}}} \\
\multicolumn{3}{|c|}{}                                                          \\ \hline
\textbf{Rank}        & \textbf{Team}             & \textbf{MSE}        \\ \hline
1                    & SIAT \cite{yang2018}                     & 0.06                         \\ \hline
2                    & VIPL\_Engagement \cite{niu2018}          & 0.07                         \\ \hline
3                    & IIIT\_Bangalore \cite{thomas2018}          & 0.08                         \\ \hline
4                    & Liulishuo \cite{chang2018}               & 0.08                         \\ \hline
5                    & Touchstone                & 0.09                         \\ \hline
6                    & \textbf{Baseline}                & 0.15                         \\ \hline
7                    & CVSP\_NTUA\_Greece        & 2.97                         \\ \hline
\end{tabular}
\caption{The Table shows the comparison of participants in the student engagement prediction sub-challenge (RMSE) on the \emph{Test} set. Note that this is the initial ranking and may change before the event.}
\label{tab:Engagement}
\end{table}

\begin{table}[]
\begin{tabular}{|l|l|c|}
\hline
\multicolumn{3}{|c|}{\multirow{2}{*}{\textbf{Group-level Emotion Recognition}}} \\
\multicolumn{3}{|c|}{}                                             \\ \hline
\textbf{Rank}  & \textbf{Team}             & \textbf{Class. Accuracy (\%)} \\ \hline
1              & UD-ECE \cite{guo2018}                    & 68.08                 \\ \hline
2              & SIAT  \cite{wang2018}                    & 67.49                 \\ \hline
3              & UofSC  \cite{khan2018}                   & 66.29                 \\ \hline
4              & LIVIA \cite{gupta2018}                    & 64.83                 \\ \hline
5              & ZJU\_CADLiu\_HanchaoLi    & 62.94                 \\ \hline
6              & ZJU\_IDI                  & 62.90                 \\ \hline
7              & FORFUN                    & 62.11                 \\ \hline
8              & SituTech                  & 61.97                 \\ \hline
9              & midea                     & 61.31                 \\ \hline
10             & \textbf{Baseline}         & 61.00                \\ \hline
11             & Beijing Normal University & 59.28                 \\ \hline
12             & UNIMIB-IVL                & 57.82                 \\ \hline
13             & AMIKAIST                  & 39.46                 \\ \hline
\end{tabular}
\caption{The Table shows the comparison of participants in the Group-level emotion recognition sub-challenge (Classification Accuracy) on the \emph{Test} set. Note that this is the initial ranking and may change before the event.}
\label{tab:Group}
\end{table}

\begin{table}[]
\begin{tabular}{|l|l|c|}
\hline
\multicolumn{3}{|c|}{\multirow{2}{*}{\textbf{Audio-Video Emotion Recognition}}} \\
\multicolumn{3}{|c|}{}                                                   \\ \hline
\textbf{Rank}    & \textbf{Team}               & \textbf{Class. Accuracy (\%)}   \\ \hline
1                & SituTech  \cite{liu2018}                 & 61.87                   \\ \hline
2                & E-HKU \cite{fan2018}                       & 61.10                   \\ \hline
3                & AIPL \cite{lu2018}                       & 60.64                   \\ \hline
3                & OL\_UC \cite{vielzeuf2018}                     & 60.64                   \\ \hline
5                & UoT                         & 60.49                   \\ \hline
6                & NLPR                        & 60.34                   \\ \hline
7                & INHA                        & 59.72                   \\ \hline
8                & SIAT                        & 58.04                   \\ \hline
9                & TsinghuaUniversity          & 57.12                   \\ \hline
10               & AIIS-LAB                    & 56.51                   \\ \hline
11               & VU                          & 56.05                   \\ \hline
12               & UofSC                       & 55.74                   \\ \hline
13               & VIPL-ICT-CAS                & 55.59                   \\ \hline
14               & Irip                        & 55.13                   \\ \hline
15               & ZBC\_Lab                    & 54.98                   \\ \hline
16               & Summerlings                 & 54.82                   \\ \hline
17               & EmoLab                      & 54.21                   \\ \hline
18               & CNU                         & 53.75                   \\ \hline
19               & Mind                        & 53.60                   \\ \hline
20               & Midea                       & 53.45                   \\ \hline
21               & KoreaUniversity             & 53.14                   \\ \hline
22               & Kaitou                      & 51.76                   \\ \hline
23               & Beijing Normal University   & 50.54                   \\ \hline
24               & USTC\_NELSLIP               & 48.70                   \\ \hline
25               & PopNow                      & 48.09                   \\ \hline
26               & BUCT                        & 45.94                   \\ \hline
27               & BIICLab                     & 42.57                   \\ \hline
28               & CobraLab                    & 41.81                   \\ \hline
29                & \textbf{Baseline}          & 41.07                   \\ \hline
30               & 17-AC                       & 35.83                   \\ \hline
31               & SAAMWILD                    & 33.84                   \\ \hline
32               & Juice                       & 25.27                   \\ \hline
\end{tabular}
\caption{The Table shows the comparison of participants in the audio-video emotion recognition sub-challenge (Classification Accuracy) on the \emph{Test} set. Note that this is the initial ranking and may change before the event.}
\label{tab:AV}
\end{table}

\section{Conclusion}
The sixth Emotion Recognition in the Wild is a challenge in
the ACM International Conference on Multimodal Interaction 2018,
Boulder. There are three sub-challenges in EmotiW 2018. Engagement detection of students while watching MOOCs is the first
sub-challenge which deals with the task of engagement recognition from the recorded videos of subjects while watching the stimuli. The second
sub-challenge is related to emotion recognition based on the
universal emotion categories from the audio-visual data collected from movies. The third sub-challenge is related to collective emotions at group-level from the images.
Different interesting methods were proposed by the challenge participants to solve these sub-challenges. The top performing methods are based on deep learning in all the sub-challenges, specifically based on ensemble of networks. 

\section{Acknowledgement}
	We are grateful to the ACM ICMI 2018 chairs, the EmotiW program committee members, the reviewers and the challenge participants. We aslso acknowledge the support of Nvidia for their generous GPU grant.
	
	\section{APPENDIX} \label{sec:Appendix}
	Movie Names: 21, 50 50, About a boy, A Case of You, After the sunset, Air Heads, American,
	American History X, And Soon Came the Darkness,
	Aviator, Black Swan, Bridesmaids, Captivity, Carrie, Change
	Up, Chernobyl Diaries, Children of Men, Contraband, Crying Game, Cursed, December
	Boys, Deep Blue Sea, Descendants, Django, Did You Hear
	About the Morgans?, Dumb and Dumberer: When Harry
	Met Lloyd, Devil's Due, Elizabeth, Empire of the Sun, Enemy at the Gates, Evil Dead, Eyes
	Wide Shut, Extremely Loud \& Incredibly Close, Feast, Four
	Weddings and a Funeral, Friends with Benefits, Frost/Nixon,
	Geordie Shore Season 1, Ghoshtship, Girl with a Pearl Earring, Gone In Sixty Seconds, Gourmet Farmer Afloat Season 2, Gourmet Farmer Afloat Season 3,
	Grudge, Grudge 2, Grudge 3, Half Light, Hall Pass, Halloween,
	Halloween Resurrection, Hangover, Harry Potter and the
	Philosopher's Stone, Harry Potter and the Chamber of Secrets,
	Harry Potter and the Deathly Hallows Part 1, Harry
	Potter and the Deathly Hallows Part 2, Harry Potter and
	the Goblet of Fire, Harry Potter and the Half Blood Prince,
	Harry Potter and the Order Of Phoenix, Harry Potter and
	the Prisoners Of Azkaban, Harold \& Kumar go to the White
	Castle, House of Wax, I Am Sam, It's Complicated, I Think
	I Love My Wife, Jaws 2, Jennifer's Body, Life is Beautiful, Little Manhattan,
	Messengers, Mama, Mission Impossible 2, Miss March,
	My Left Foot, Nothing but the Truth, Notting Hill, Not Suitable for Children, One
	Flew Over the Cuckoo's Nest, Orange and Sunshine, Orphan, Pretty
	in Pink, Pretty Woman, Pulse, Rapture Palooza, Remember Me, Runaway Bride, Quartet, Romeo Juliet, Saw 3D, Serendipity, Silver Lining
	Playbook, Solitary Man, Something Borrowed, Step Up 4, Taking Lives, Terms
	of Endearment, The American, The Aviator, The Big Bang Theory, The Caller, The Crow,
	The Devil Wears Prada, The Eye, The Fourth Kind, The Girl with Dragon Tattoo, The
	Hangover, The Haunting, The Haunting of Molly Hartley, The Hills have Eyes 2, The Informant!,
	The King's Speech, The Last King of Scotland, The Pink Panther 2, The Ring 2, The Shinning, The
	Social Network, The Terminal, The Theory of Everything, The Town, Valentine Day,
	Unstoppable, Uninvited, Valkyrie, Vanilla Sky, Woman In
	Black, Wrong Turn 3, Wuthering Heights, You're Next, You've Got Mail.

\bibliographystyle{ACM-Reference-Format}
\balance
\bibliography{abhinavdhall}

\end{document}